\documentclass[10pt,twocolumn,letterpaper]{article}

\usepackage{iccv}
\usepackage{times}
\usepackage{epsfig}
\usepackage{graphicx}
\usepackage{amsmath}
\usepackage{amssymb}
\usepackage{pifont}
\usepackage{fontawesome5}

\usepackage{sg-macros}
\usepackage{tabularx} 
\newcolumntype{C}{>{\centering\arraybackslash}X}
\usepackage{booktabs}
\makeatletter\let\captiontemp\@makecaption\makeatother
\usepackage[font=small]{subcaption}
\makeatletter\let\@makecaption\captiontemp\makeatother

\makeatletter
\def\@fnsymbol#1{\ensuremath{\ifcase#1\or \dagger\or \ddagger\or
\mathsection\or \mathparagraph\or \|\or **\or \dagger\dagger
\or \ddagger\ddagger \else\@ctrerr\fi}}
\makeatother

\newcommand\codeurl[1]{{{\color{blue}{\url{#1}}}}}

\usepackage[pagebackref=true,breaklinks=true,letterpaper=true,colorlinks,bookmarks=false]{hyperref}
\usepackage[capitalize]{cleveref} 
\crefname{section}{Sec.}{Secs.}
\Crefname{section}{Section}{Sections}
\Crefname{table}{Table}{Tables}
\crefname{table}{Tab.}{Tabs.}
\Crefname{figure}{Figure}{Figures}
\crefname{figure}{Fig.}{Figs.}
\Crefname{equation}{Equation}{Equations}
\crefname{equation}{eq.}{eqs.}

\iccvfinalcopy 

\ificcvfinal\fi

\begin{document}

\title{Exploring Predicate Visual Context in Detecting Human--Object Interactions}

\author{Frederic Z. Zhang$^{1}$\thanks{Work done at Microsoft Research Asia.} \quad Yuhui Yuan$^{2}$ \quad Dylan Campbell$^{1}$ \quad Zhuoyao Zhong$^{2}$ \quad Stephen Gould$^{1}$ \\
$^1$The Australian National University \quad $^2$Microsoft Research Asia\\
{\tt\small frederic.zhang@anu.edu.au \quad yuhui.yuan@microsoft.com} \\
{\faIcon{github} \small \codeurl{{https://github.com/fredzzhang/pvic}}}
}

\maketitle
\ificcvfinal\thispagestyle{empty}\fi

\begin{abstract}
Recently, the DETR framework has emerged as the dominant approach for human--object interaction (HOI) research.
In particular, two-stage transformer-based HOI detectors are amongst the most performant and training-efficient approaches. However, these often condition HOI classification on object features that lack fine-grained contextual information, eschewing pose and orientation information in favour of visual cues about object identity and box extremities.
This naturally hinders the recognition of complex or ambiguous interactions. In this work, we study these issues through visualisations and carefully designed experiments. Accordingly, we investigate how best to re-introduce image features via cross-attention.
With an improved query design, extensive exploration of keys and values, and box pair positional embeddings as spatial guidance, our model with enhanced  \textbf{p}redicate \textbf{vi}sual \textbf{c}ontext (PViC) outperforms state-of-the-art methods on the HICO-DET and V-COCO benchmarks, while maintaining low training cost.
\end{abstract}

\section{Introduction}

\begin{figure}[t]
   \begin{subfigure}[t]{0.49\linewidth}
      \centering
      \includegraphics[width=\linewidth]{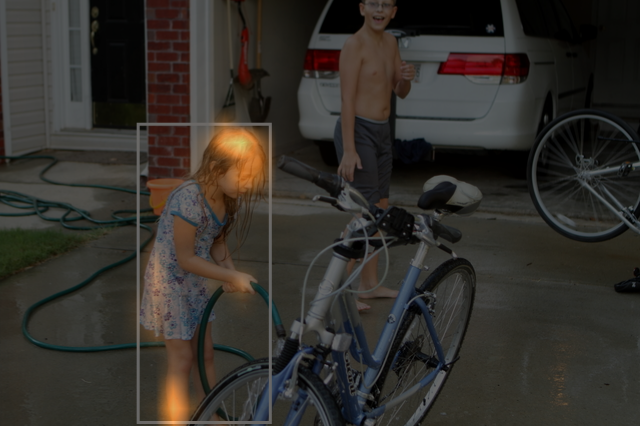}
      \caption{Attn. for human (DETR)}
      \label{fig:teaser-hum-attn}
   \end{subfigure}%
   \hfill%
   \begin{subfigure}[t]{0.49\linewidth}
      \centering
      \includegraphics[width=\linewidth]{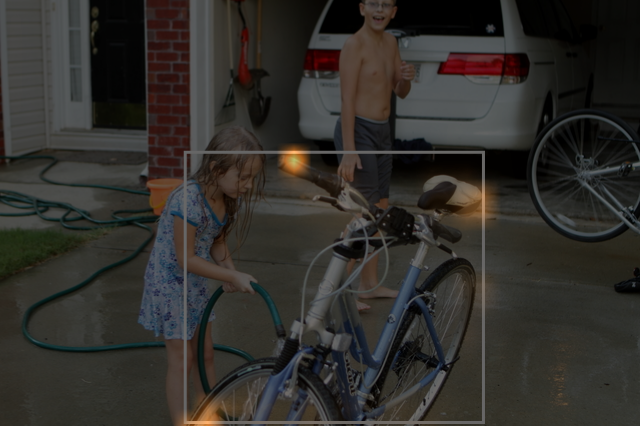}
      \caption{Attn. for bike (DETR)}
      \label{fig:teaser-bike-attn}
   \end{subfigure}%
	\vfill%
   \begin{subfigure}[t]{0.49\linewidth}
      \centering
      \includegraphics[width=\linewidth]{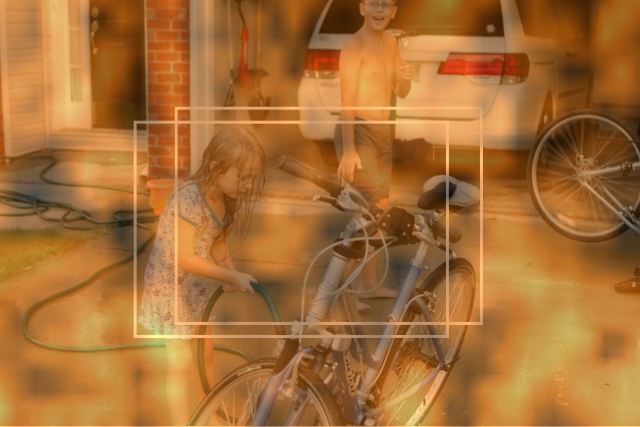}
      \caption{Attn. for triplet (QPIC)}
      \label{fig:teaser-qpic-attn}
   \end{subfigure}%
   \hfill%
   \begin{subfigure}[t]{0.49\linewidth}
      \centering
      \includegraphics[width=\linewidth]{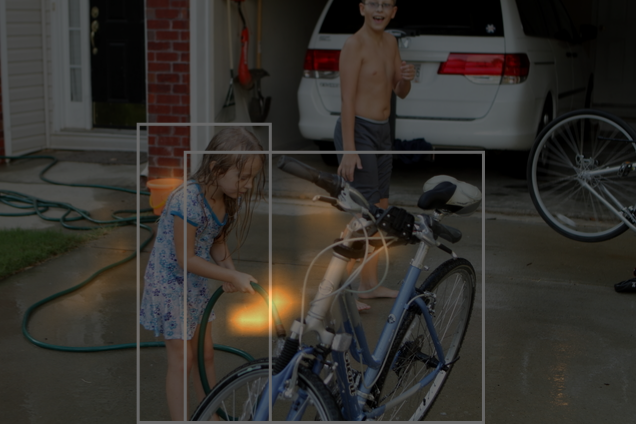}
      \caption{Attn. for predicate (Ours)}
      \label{fig:teaser-ours-attn}
   \end{subfigure}%
   \vspace{10px}
   \caption{Visual context for the (\subref{fig:teaser-hum-attn}, \subref{fig:teaser-bike-attn}) two-stage HOI detector UPT~\cite{upt}, (\subref{fig:teaser-qpic-attn}) one-stage HOI detector QPIC~\cite{qpic} and (\subref{fig:teaser-ours-attn}) our method. Cross-attention weights from the last decoder layer are used for visualization. UPT uses coarse object features that favor visual cues about object identity and box extremities. QPIC fails to detect the triplet \textit{person-washing-bike} as it struggles to locate the relevant visual context (person, bike, and the water hose). The box pair with the highest IoUs against the ground truth is selected for display. Our two-stage method with pre-detected objects successfully recognises the predicate \textit{washing} as it pinpoints the location of the image region containing the water hose.}
   \label{fig:teaser}
\end{figure}

Detecting human--object interactions (HOI) is the task of localising and recognising interactive human--object pairs. It extends the detection of objects to include their relationships and facilitates a deeper understanding of visual scenes. 
Recent developments in the detection of human--object interactions have largely adhered to the encoder--decoder style introduced by the detection transformers (DETR)~\cite{detr}, where learnable queries are randomly initialised with Gaussian noise, and progressively decoded into the desired \textit{human–predicate–object} triplets. Such one-stage detectors~\cite{qpic, hotr, hoitrans, asnet, cdn, ssrt} require pre-trained DETR weights for initialisation to facilitate stable convergence. As we will demonstrate empirically, the pre-trained encoder features have overfitted to object cues and lack the necessary information for recognising human--object interactions. This means that the transformer encoder weights need to change significantly to produce discriminative features for such tasks. Together with the need to repurpose the decoder to detect HOI triplets rather than unary objects, this results in long training schedules that often amount to hundreds of GPU hours. On the other hand, two-stage detectors adopt a different methodology, wherein an object detector is fine-tuned and then frozen. These approaches focus on the extraction and exploitation of the rich information residing in the frozen detector. Naturally, two-stage detectors require significantly less time and resources to train, facilitating more model analysis and experimentation.

\begin{figure*}[t]
   \begin{subfigure}[t]{0.49\linewidth}
      \centering
      \includegraphics[height=100px]{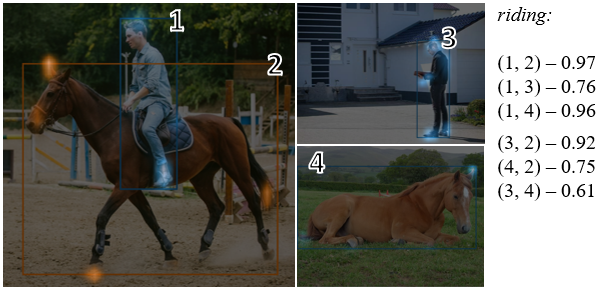}
      \caption{Object feature replacement for predicate \textit{riding}.}
      \label{fig:obj_feat_riding}
   \end{subfigure}
   \begin{subfigure}[t]{0.49\linewidth}
      \centering
      \includegraphics[height=100px]{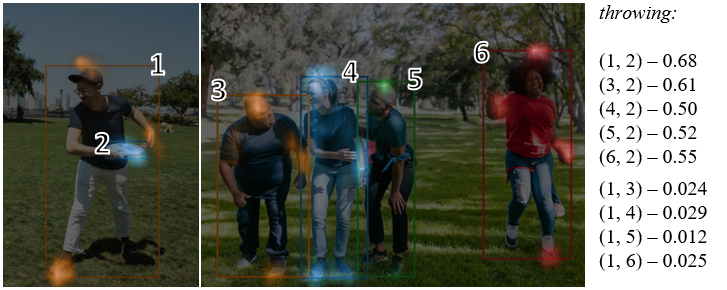}
      \caption{Object feature replacement for predicate \textit{throwing}.}
      \label{fig:obj_feat_throwing}
   \end{subfigure}
   \vspace{5px}
   \caption{
   	Object features from a frozen object detector (DETR~\cite{detr}) are extracted from image regions that are indicative of the object identity and are near the bounding box extremities, as shown by the cross-attention weights overlaid on each detected object. These features often lack the fine-grained information to recognise HOIs. As a result, replacing object features with those from an object in a different~(\subref{fig:obj_feat_riding},~\subref{fig:obj_feat_throwing}) pose, orientation or even~(\subref{fig:obj_feat_riding}) identity does not impact the classification score significantly. Experiments are conducted on UPT~\cite{upt}, where the spatial configuration for pair (1, 2) is used for all pairs in each set of images while the object features are replaced.
   }
   \label{fig:obj_feat}
\end{figure*}

Current state-of-the-art two-stage detector UPT~\cite{upt} employs a fine-tuned DETR detector, and performs self-attention on unary (object) and pairwise (human--object) tokens. Despite its overall high performance and low cost, it only utilises object features from the frozen detector, complemented by hand-crafted spatial features, to construct the final representations. As we show in Figures~\ref{fig:teaser-hum-attn} and \ref{fig:teaser-bike-attn}, these frozen features are obtained by attending to image regions indicative of object identity and box extremities, thus lacking the necessary information for recognising HOIs.

\begin{figure*}[t]\captionsetup[subfigure]{font=footnotesize}
   \begin{subfigure}[t]{0.245\linewidth}
     \centering
     \includegraphics[width=\linewidth]{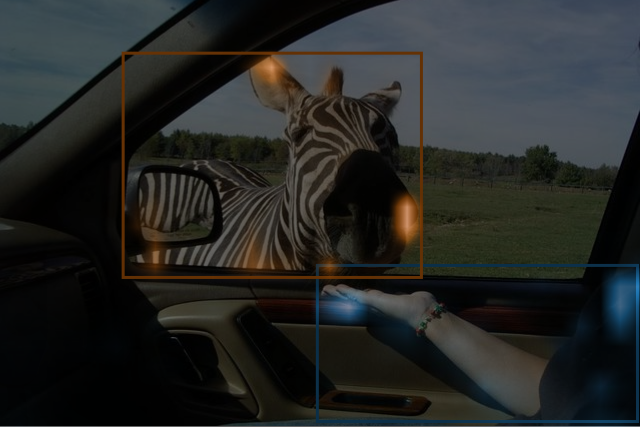}
     \caption{Feeding. UPT: $0.07$}
     \label{fig:comp_1}
   \end{subfigure}%
   \hfill
   \begin{subfigure}[t]{0.245\linewidth}
     \centering
     \includegraphics[width=\linewidth]{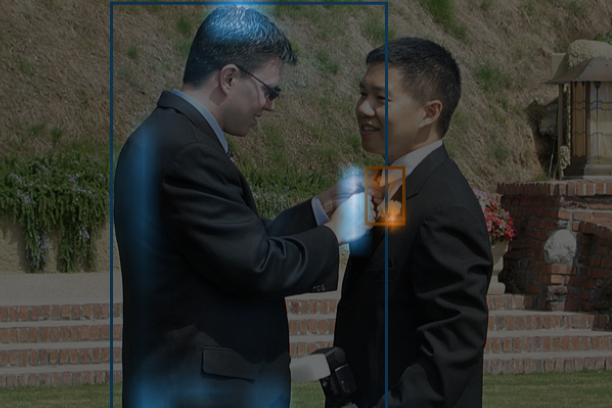}
     \caption{Tying. UPT: $0.12$}
     \label{fig:comp_3}
   \end{subfigure}%
   \hfill
   \begin{subfigure}[t]{0.245\linewidth}
     \centering
     \includegraphics[width=\linewidth]{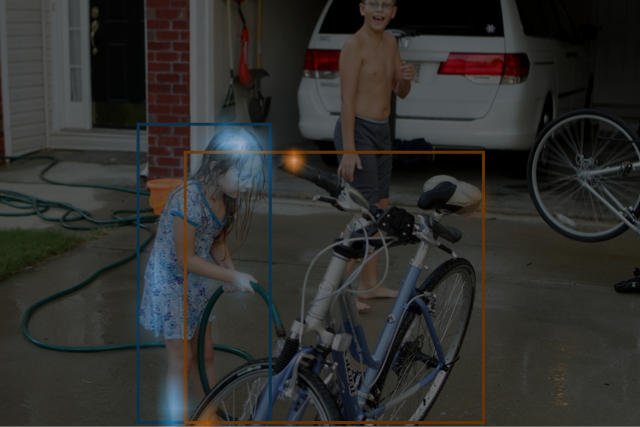}
     \caption{Washing. UPT: $0.12$}
     \label{fig:comp_2}
   \end{subfigure}%
   \hfill
   \begin{subfigure}[t]{0.245\linewidth}
     \centering
     \includegraphics[width=\linewidth]{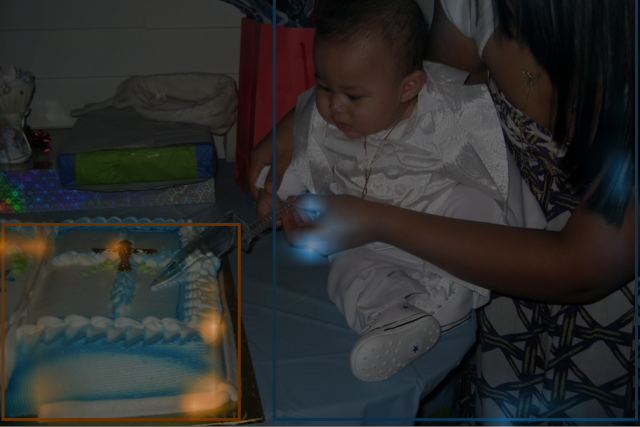}
     \caption{Cutting. UPT: $0.10$}
     \label{fig:comp_4}
   \end{subfigure}%
   \vfill%
   \begin{subfigure}[t]{0.245\linewidth}
      \centering
      \includegraphics[width=\linewidth]{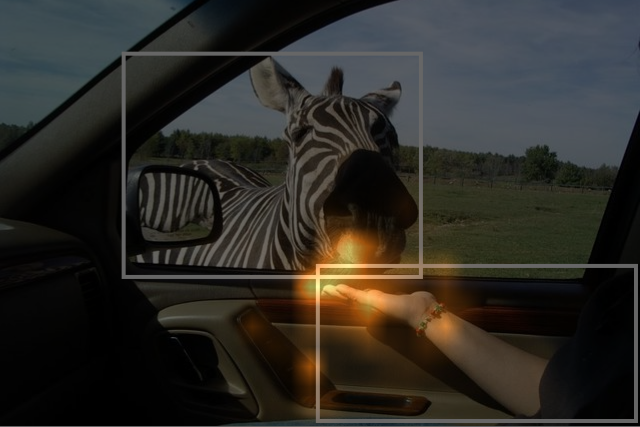}
      \caption{Feeding. Ours: \textbf{0.75}}
      \label{fig:comp_1_ours}
   \end{subfigure}%
   \hfill
   \begin{subfigure}[t]{0.245\linewidth}
      \includegraphics[width=\linewidth]{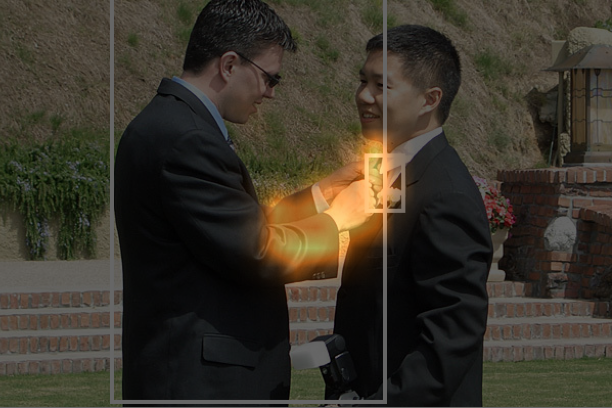}
      \caption{Tying. Ours: \textbf{0.52}}
      \label{fig:comp_3_ours}
   \end{subfigure}%
   \hfill
   \begin{subfigure}[t]{0.245\linewidth}
      \centering
      \includegraphics[width=\linewidth]{figures/attn_concat.png}
      \caption{Washing. Ours: \textbf{0.66}}
      \label{fig:comp_2_ours}
   \end{subfigure}%
   \hfill
   \begin{subfigure}[t]{0.245\linewidth}
      \includegraphics[width=\linewidth]{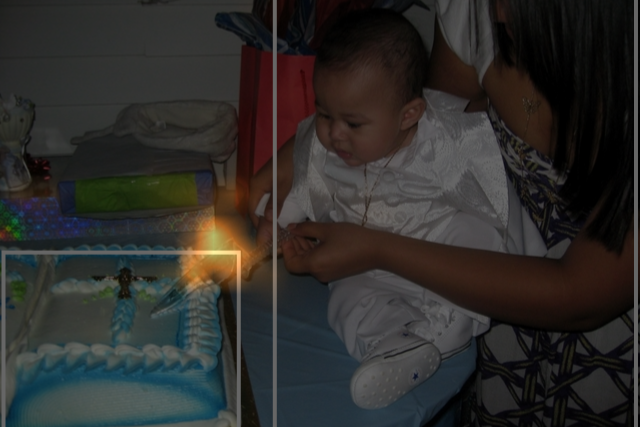}
      \caption{Cutting. Ours: \textbf{0.61}}
      \label{fig:comp_4_ours}
   \end{subfigure}%
   \vfill%
   \begin{subfigure}[t]{0.245\linewidth}
      \centering
      \includegraphics[width=\linewidth]{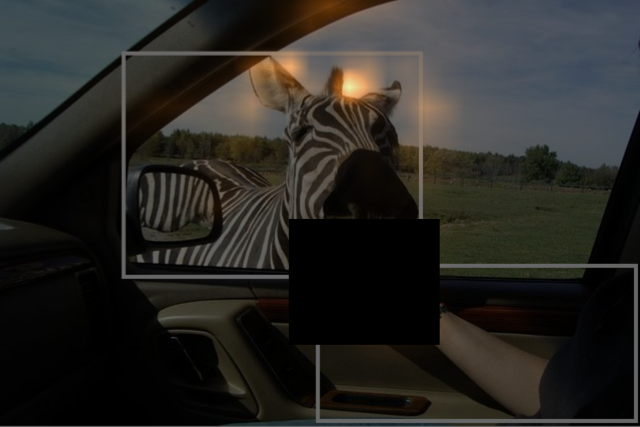}
      \caption{Feeding. Ours (masked): 0.09}
      \label{fig:comp_1_blocked_ours}
   \end{subfigure}%
   \hfill
   \begin{subfigure}[t]{0.245\linewidth}
      \includegraphics[width=\linewidth]{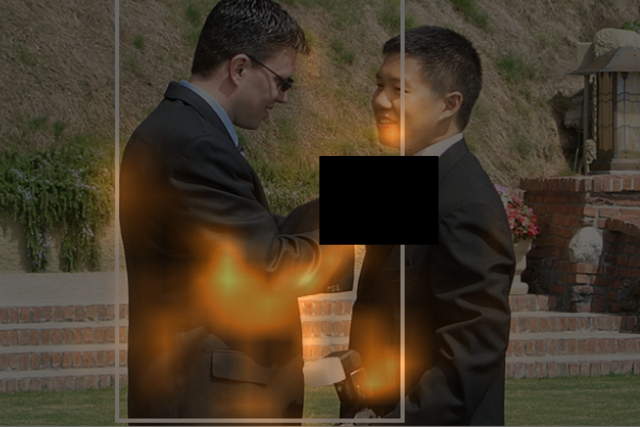}
      \caption{Tying. Ours (masked): 0.14}
      \label{fig:comp_3_blocked_ours}
   \end{subfigure}%
   \hfill
   \begin{subfigure}[t]{0.245\linewidth}
      \centering
      \includegraphics[width=\linewidth]{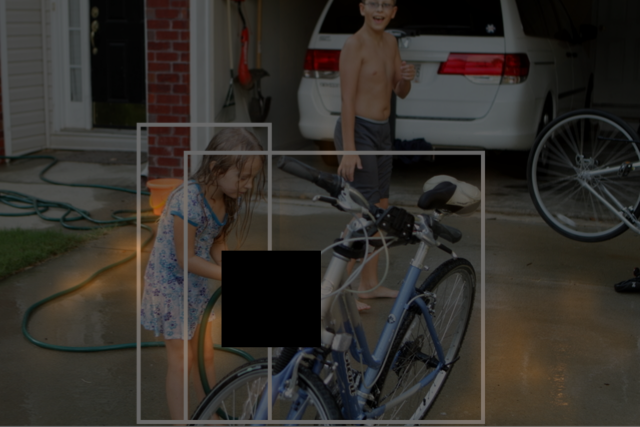}
      \caption{Washing. Ours (masked): 0.21}
      \label{fig:comp_2_blocked_ours}
   \end{subfigure}%
   \hfill
   \begin{subfigure}[t]{0.245\linewidth}
      \includegraphics[width=\linewidth]{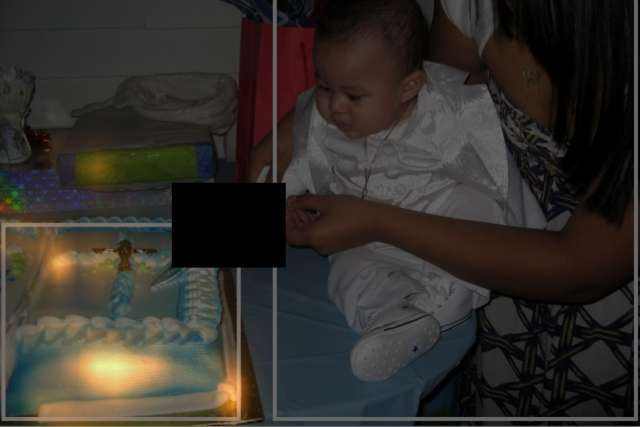}
      \caption{Cutting. Ours (masked): 0.08}
      \label{fig:comp_4_blocked_ours}
   \end{subfigure}%
   \vspace{5px}
   \caption{
   	Existing two-stage HOI detectors (\eg, UPT~\cite{upt}) lack relevant visual context, including (\subref{fig:comp_1},~\subref{fig:comp_3}) fine-grained information about the subject or object, such as human pose, and (\subref{fig:comp_2},~\subref{fig:comp_4}) other relevant contextual information in the scene, such as another object involved in the interaction. The predicted score for each example is listed in the caption. UPT (first row) uses frozen object features which often pool information from the box boundary since this aids localisation. Consequently, such features do not cover other aspects of the object and are not discriminative enough to recognise complex human--object interactions. Our method (second row) solves such failure cases with spatially guided cross-attention, pinpointing the image regions corresponding to the relevant body parts or the additional object besides the human--object pair. To demonstrate that these regions are indeed highly relevant to the prediction score, we mask out those image regions with the highest attention weights (third row), and observe a significant drop in prediction scores.}
   \label{fig:comp}
\end{figure*}

In \Cref{fig:obj_feat}, we study the impacts of this lack of information by replacing the object features with those of a different object.
For predicates with a distinct spatial pattern, such as \textit{riding}, we observe that the predicted scores do not change significantly when object features are swapped, suggesting that the spatial information dominates the visual information, as shown in \Cref{fig:obj_feat_riding}.
Even when the replacement features come from objects of a different identity, UPT still gives confident predictions for the same predicate, such as (4, 2) \textit{horse--riding--horse}. Yet, the majority of predicates do not exhibit a prominent spatial pattern, \eg \textit{throwing}. In such cases, visual context plays a crucial role. Naturally, replacing object features amounts to a more tangible impact (\Cref{fig:obj_feat_throwing}). However, the score drop when replacing the human features with those of a non-interactive person is still not significant, indicating that such features do not contain enough visual cues to differentiate interactiveness. As we will show in~\Cref{fig:comp}, failure cases of UPT often require much richer visual context. In particular, we identify two types of context that coarse object features lack: 
fine-grained information about the subject or object, such as human pose, and information about other relevant context in the scene, such as another object involved in the interaction.

To address the aforementioned issues, we investigate how to enrich the contextual cues for human--object pair representations. Our contribution is twofold. We conduct thorough analysis with abundant visualisations to characterise the two types of visual contexts lacking in current two-stage models and the damage this causes. Accordingly, we develop a superior two-stage detector with a lightweight decoder, where we improve the query design with a more streamlined architecture, explore various choices and compositions of keys/values and introduce positional embeddings tailored for bounding box pairs. In particular, we demonstrate that the positional embeddings function as spatial guidance in cross-attention, and shed light on this mechanism with rich visualisations.

\section{Related Works}

There is a large body of works~\cite{qpic, hotr, hoitrans, asnet} centred around adapting the detection transformer~\cite{detr} to one-stage HOI detectors. Since Tamura \etal~\cite{qpic} established a strong baseline, the focus has shifted to improving the architecture design. Zhang \etal~\cite{cdn} proposed to partially decouple the feature representation of humans and objects from that of the predicates. 
Qu \etal~\cite{doq} investigated ways to better utilise the ground truth with data distillation. 
Tu \etal~\cite{iwin} and Kim \etal~\cite{mstr} explored multiscale backbone features by either exploiting irregular window attention or extending Deformable DETR~\cite{deform-detr} to HOI detection. Last, Wu \etal~\cite{partmap} demonstrated the value of human pose by applying body-part masks in transformer cross-attention.

Two-stage detection has received much less attention in comparison. Graph-based methods~\cite{scg, drg, vsgnet} were the state of the art for an extended period of time. Since the advent of transformer-based approaches, much of the focus has been shifted to one-stage detection. Recently, Zhang \etal~\cite{upt} demonstrated that self-attention can be repeatedly applied to unary objects and human--object pairs, achieving complementary effects. However, the lack of contextual information is its major weakness. A concurrent work~\cite{stip} addressed this by integrating hand-coded HOI structures into transformer cross-attention. Nonetheless, this introduces even more hand-crafted elements into two-stage detection. Seeking a more streamlined model design, we show that a dedicated query positional embedding yields better performance and more interpretable visualisations.

In addition, there have been some works focusing on other aspects of HOI research. Specifically, Wang \etal~\cite{odm} studied the object bias and explored ways to mitigate it. Yuan \etal~\cite{rlip} conducted contrastive language--image pre-training for HOI representations and demonstrated its effectiveness. Liao \etal~\cite{gen-vlkt} explored data distillation from CLIP~\cite{clip} features and showed competitive performance.

\section{Spatially Guided Cross-Attention}

\begin{figure}[t]\captionsetup[subfigure]{font=footnotesize}
   \begin{subfigure}[t]{0.48\linewidth}
     \centering
     \includegraphics[width=\linewidth]{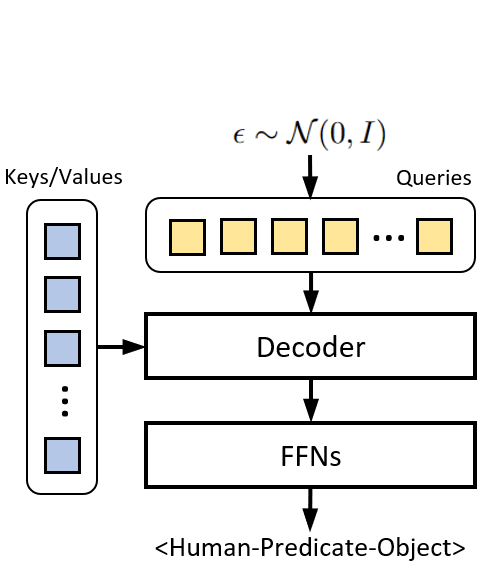}
     \caption{Implicit Queries}
     \label{fig:imp_q}
   \end{subfigure}
   \begin{subfigure}[t]{0.48\linewidth}
   \centering
     \includegraphics[width=\linewidth]{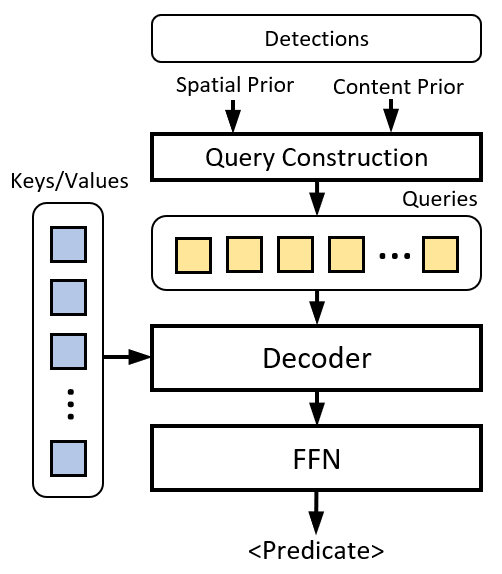}
     \caption{Explicit Queries}
     \label{fig:exp_q}
   \end{subfigure}
   \vspace{5px}
   \caption{Implicit queries~(\subref{fig:imp_q}) for one-stage detectors and explicit queries~(\subref{fig:exp_q}) for two-stage detectors.}
   \label{fig:decoder}
\end{figure}

The underpinning of query-based detection systems is the transformer cross-attention mechanism~\cite{xfmer}, which acts as a form of soft RoI pooling where the weights are computed dynamically from the data. Stacking cross-attention layers allows the queries to aggregate useful information from the keys/values (image features) gradually. In the detection transformer, queries are randomly initialised with Gaussian noise and learn to represent spatial priors (box centre positions, widths and heights)~\cite{detr} as training progresses. We refer to such queries as \textit{implicit queries} (\Cref{fig:imp_q}), commonly used in one-stage HOI detectors. For their two-stage counterparts~\cite{upt, stip}, thanks to the rich information in the detections, there is no need for such learned queries. Instead, the queries are explicit human--object pair representations, injected with spatial and content priors. We refer to them as \textit{explicit queries} (\Cref{fig:exp_q}).

\subsection{Explicit Queries}

Prior to the query construction, we filter the detected objects by their scores and perform self-attention to refine the object features. As Zhang \etal~\cite{upt} pointed out, such self-attention promotes information flow between the interactive objects and helps increase scores of positive examples. Based on the observation that interactive instances tend to appear close together in an image, we apply positional embeddings for bounding boxes, which encourages attention between near objects. Insights on this design will be detailed in the next section. 
Formally, denote the mapping from a scalar to a sinusoidal embedding by $\boldsymbol{\phi}: \mathbb{R} \to \mathbb{R}^d$,
\begin{equation}
   \boldsymbol{\phi}(x)_{2i} = \sin\left(\frac{x}{\tau^{2i/d}}\right),\; \boldsymbol{\phi}(x)_{2i - 1} = \cos\left(\frac{x}{\tau^{2i/d}}\right),
\end{equation}
where $i=1,\ldots,d/2$ and $\tau$ is a temperature parameter. A bounding box can be encoded by concatenating the sinusoidal embeddings of the centre coordinates, width and height, which are then applied via element-wise sum as a convention~\cite{xfmer, detr}. We show in the experiment section that with the positional embeddings, vanilla self-attention achieves better performance than the custom layer in UPT.

\begin{figure}[t]
   \centering
   \includegraphics[width=.7\linewidth]{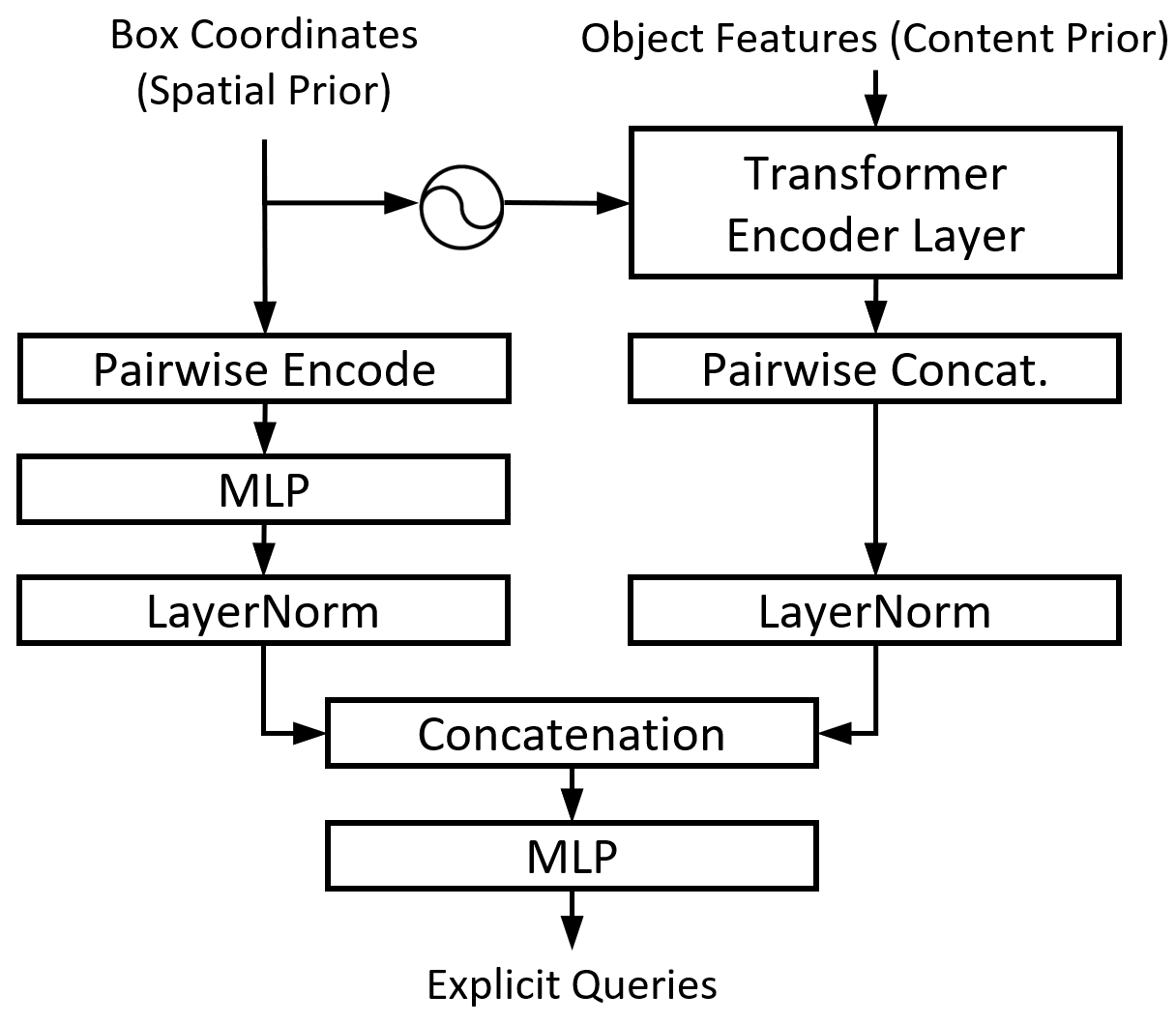}
   \caption{An illustration of the construction of explicit queries.}
   \label{fig:eq-construction}
\end{figure}

To construct explicit queries, we enumerate all human--object pairs. For each pair, the representation is obtained by fusing the concatenated object features and their spatial representations, following previous practice~\cite{scg, upt}. In addition, we apply LayerNorm~\cite{ln} to both modalities before fusion. This greatly stabilises the training process and prevents numeric overflow, which was previously resolved by using large batch sizes. The full process of query construction is illustrated in~\Cref{fig:eq-construction}.

\subsection{Positional Embeddings as Guidance}

\textbf{Cross-Attention}.
Even though the explicit query representation of the human--object pair already contains spatial priors, positional embeddings are still critical since they function as spatial biases on the attention weights. This is particularly important in the case of cross-attention. To shed light on its impact, let us denote the keys and queries as $\bk_c$ and $\bq_c$, and their respective positional embeddings as $\bk_p$ and $\bq_p$. For simplicity, let us omit the linear transformations and the normalisation. Dot-product attention is computed as
\begin{equation}
   (\bk_c + \bk_p)\transpose (\bq_c + \bq_p) = {\bk_c}\transpose\bq_c + \ldots + {\bk_p}\transpose\bq_p.
   \label{eq:dot_p}
\end{equation}
Intuitively, first term on the RHS measures the similarity between the content features of the keys (image features) and queries, while the last term measures that of the positional embeddings. More specifically, for an image token with normalised spatial indices $(i, j)$ and a 2D point with normalised coordinate $(x, y)$, the last term can be expanded as a simple sum of similarity between coordinates,
\begin{equation}
   {\bk_p}\transpose\bq_p = \boldsymbol{\phi}(i)\transpose \boldsymbol{\phi}(x) + \boldsymbol{\phi}(j)\transpose \boldsymbol{\phi}(y).
\end{equation}
As an advantage of explicit queries, the availability of box coordinates allows us to use box centres to construct positional embeddings, directly adding a bias to the attention map in the corresponding position. A weakness of the aforementioned positional embeddings is the lack of information on box dimensions. Although the subsequent linear transformations have the potential to shift and deform the dot-product attention, Liu \etal~\cite{dab-detr} showed that the positional embeddings can be modulated with box widths and heights, saving the network from learning the relevant transforms. For a bounding box $\bb = [x, y, w, h]$, we follow their practice by using the normalised widths and heights as different temperature parameters in horizontal and vertical directions for the subsequent softmax normalisation, leading to a bias term on attention weights as below
\begin{equation}
   {\bk_p}\transpose\bq_p = \boldsymbol{\phi}(i)\transpose \boldsymbol{\phi}(x) \frac{w_\text{ref}}{w} + \boldsymbol{\phi}(j)\transpose \boldsymbol{\phi}(y) \frac{h_\text{ref}}{h},
   \label{eq:mod_attn}
\end{equation}
where $w_\text{ref}$ and $h_\text{ref}$ are reference values learned from the box features using a two-layer MLP, as follows
\begin{equation}
   w_\text{ref}, h_\text{ref} = \sigma(\text{MLP}(\boldf)),
\end{equation}
with $\sigma$ being the sigmoid function and $\boldf$ being the box appearance features obtained from the object detector.

\begin{figure}[t]
   \begin{subfigure}[t]{.24\linewidth}
      \centering
      \includegraphics[width=\linewidth]{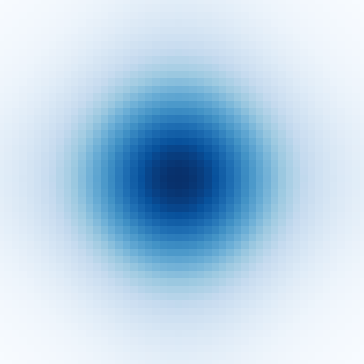}
      \caption{(0.5, 0.5)}
      \label{fig:mod_attn_1}
   \end{subfigure}
   \hfill
   \begin{subfigure}[t]{.24\linewidth}
      \centering
      \includegraphics[width=\linewidth]{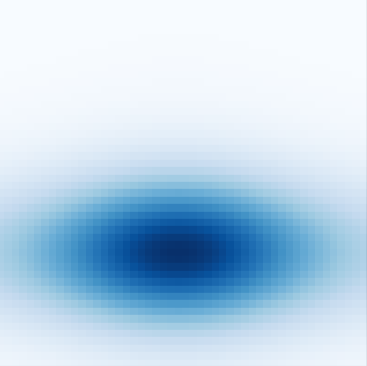}
      \caption{(0.5, 0.7)}
      \label{fig:mod_attn_2}
   \end{subfigure}
   \hfill
   \begin{subfigure}[t]{.24\linewidth}
      \centering
      \includegraphics[width=\linewidth]{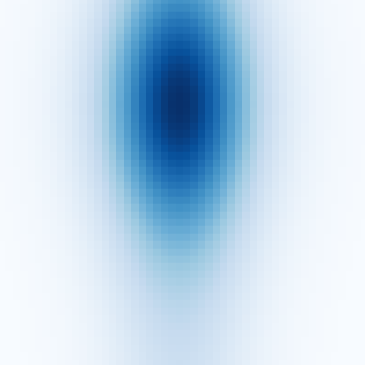}
      \caption{(0.5, 0.3)}
      \label{fig:mod_attn_3}
   \end{subfigure}
   \hfill
   \begin{subfigure}[t]{.24\linewidth}
      \centering
      \includegraphics[width=\linewidth]{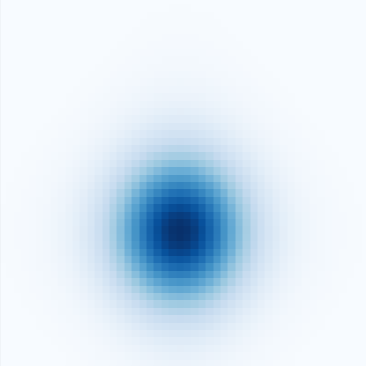}
      \caption{(\subref{fig:mod_attn_2}, \subref{fig:mod_attn_3})}
      \label{fig:mod_attn_4}
   \end{subfigure}
   \vspace{5px}
   \caption{Visualisation of the dot-product attention between the position embeddings of 2D points and those of image patches, without height and width modulation~(\subref{fig:mod_attn_1}), with modulation (\subref{fig:mod_attn_2}, \subref{fig:mod_attn_3}). Point coordinates are listed in the captions. The spatially summed attention weights \textit{after} softmax normalisation is shown in~(\subref{fig:mod_attn_4}).}
   \label{fig:mod_attn}
\end{figure}

To extend the usage to bounding box pairs, we concatenate the positional embeddings of the two box centres. The concatenation of two positional embeddings is equivalent to spatially summing the attention weights (Eq.~\ref{eq:mod_attn}) of two boxes prior to the softmax normalisation. We defer the mathematical details to the supplementary materials, but show visualisations in~\Cref{fig:mod_attn}. Furthermore, we show in the experiment section that for cross-attention, the two omitted cross-terms in Eq.~\ref{eq:dot_p} produce mostly noise, hampering the effectiveness of positional embeddings. Thus, we follow Meng \etal~\cite{cond-detr} to concatenate the keys and queries with their respective positional embeddings, essentially removing said terms. In addition, separate linear layers are employed for the content features (keys and queries) and positional embeddings to avoid undesired information flow, in the same spirit as removing the two cross-terms. We show visualisations on the impacts of the positional embeddings in the experiment section~(\Cref{fig:attn}).

\textbf{Self-Attention}. For explicit queries, self-attention acts mainly as a form of suppression \cite{upt}. Interactive human--object pairs, which are often the most salient ones, suppress the non-interactive pairs via the attention mechanism. The positional embeddings, on the other hand, add an inductive bias such that box pairs in close proximity attend to each other more. While such inductive bias is intuitive in cross-attention, it does not reflect the way human--object pairs interact with each other. As we did not observe any improvement, we do not use positional embeddings in self-attention between human--object pairs. In contrast, self-attention amongst the unary objects does benefit from positional embeddings, because interactive objects tend to appear together, and often share an intersected area. Thus, such an inductive bias promotes attention between near instances and aids the training process.

\subsection{Keys/Values}

In one-stage methods, the encoder features serve as dedicated keys/values and are end-to-end trained. Two-stage methods, on the other hand, employ pre-trained object detectors, mostly with frozen weights to ensure the performance of the detector. Although an additional feature head can be used for refinement, the source of the keys/values is of utmost importance. We empirically found (see \Cref{sec:ablation}) that backbone ResNet~\cite{resnet} C5 features are the most informative, and a very lightweight feature head with window attention~\cite{swint} improves the performance further, while higher feature resolutions and multiscale features do not introduce additional benefits.

\subsection{Training and Inference}

During training, we use the focal loss~\cite{retinanet} on the predicted action logits following previous practice~\cite{upt, stip}, where invalid actions for each object are masked out. During inference, we combine the object detection scores ($s_h, s_o$) and action prediction scores ($\bs_a$) using the geometric mean with hyperparameter $\lambda \in [0, 1]$
as follows
\begin{equation}
   \bs = (s_h s_o)^{1 - \lambda} \bs_a^{\lambda}.
   \label{eq:score}
\end{equation}

\section{Experiments}
\label{sec:experiments}

In this section, we first present a thorough ablation study by progressively building up the proposed model, demonstrating the impact of each design choice. We then compare our method against state-of-the-art models and show its superior performance, even against methods that perform data distillation on large pre-trained vision and language models. Last, to shed light on how spatial priors are used to guide cross-attention, we show visualisations of the attention weights for different terms in Eq.~\ref{eq:dot_p} and demonstrate why concatenated positional embeddings are superior.

\paragraph{Datasets:}
The primary dataset used for model design and validation is HICO-DET~\cite{hicodet}, which contains $37\,633$ training images and $9\,546$ test images. The dataset includes the same $80$ object classes as in MS COCO~\cite{coco}, $117$ action classes and $600$ interaction classes. For legacy reasons we also report on V-COCO~\cite{vcoco}, a much smaller dataset with $2\,533$ training images, $2\,867$ validation images and $4\,946$ test images. The dataset has $24$ action classes.

\subsection{Implementation Details}

We use fine-tuned DETR provided by Zhang \etal~\cite{upt} and freeze the weights. In addition, we fine-tuned deformable DETR~\cite{deform-detr} with iterative box refinement and the two-stage options. During training, we adopt the same sampling scheme in UPT, by filtering detections with a threshold of 0.05 and sampling a minimum of 3 and a maximum of 15 human and object instances each. For focal loss, we use $\alpha = 0.5$ and $\gamma=0.1$. The hyper-parameter $\lambda$ in the geometric mean is set to be 0.26, which is simply a normalised value and has an equivalent effect as the setup in UPT. For the feature head, we use one encoder layer with window attention and a window size of 8$\times$8. For the decoder, we use two layers.
We apply the same data augmentation in previous works~\cite{detr, qpic, upt}, including multiscale resizing, random cropping and random colour jittering. AdamW~\cite{adamw} is used as the optimiser, with both the learning rate and weight decay being $10^{-4}$. Unless otherwise specified, all models are trained for 30 epochs, with a learning rate drop by a factor of 5 at the $20^{\text{th}}$ epoch. Training is conducted on 8 Nvidia Tesla V100 GPUs, with a batch size of 16.

\subsection{Ablation Study}
\label{sec:ablation}

\begin{table}[t]\small
	\caption{The mAP ($\times$100) of model variants with different components of the decoder on the HICO-DET test set. Variant C is equivalent to UPT \cite{upt}. Results are averaged across three runs.
	}
	\label{tab:kv}
	\setlength{\tabcolsep}{2pt} 
	\begin{tabularx}{\linewidth}{@{\extracolsep{\fill}} l ccc l ccc}
		\toprule
		 & \multicolumn{3}{c}{\textbf{Decoder}} & & \multicolumn{3}{c}{\textbf{Default Setting (mAP)}} \\
      \cline{2-4} \cline{6-8} \\ [-10pt]
      \# & Self & Cross & FFN & C.A. src.  & Full & Rare & N-rare \\
		\midrule
		A & & & & None & 30.71 & 25.16 & 32.37 \\
      B & & & \checkmark & None & 30.98 & 25.36 & 32.62 \\
      C & \checkmark & & \checkmark & None & 31.47 & 25.98 & 33.11 \\
      D & \checkmark & \checkmark & \checkmark & Encoder & 31.51 & 26.12 & 33.13 \\
      E & \checkmark & \checkmark & \checkmark & Backbone & \textbf{32.89} & \textbf{27.91} & \textbf{34.38} \\
      \bottomrule
	\end{tabularx}
\end{table}

\begin{figure}[t]
   \centering
   \includegraphics[width=.6\linewidth]{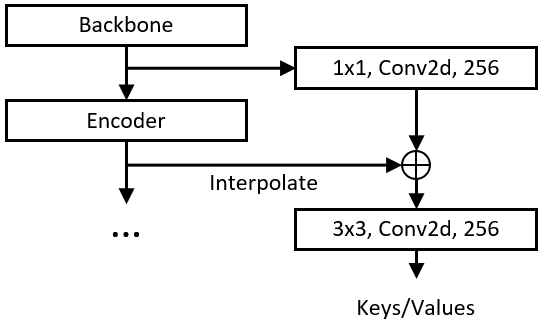}
   \caption{Illustration of the composition of backbone and encoder features. Encoder features are upsampled with bilinear interpolation when there is a resolution discrepancy.}
   \label{fig:feat_comp}
\end{figure}

\begin{table}[t]\small
	\caption{The mAP ($\times$100) of model variants with different choices and compositions of features as cross-attention keys/values on the HICO-DET test set. Results are averaged across three runs.}
	\label{tab:kv_comp}
	\setlength{\tabcolsep}{4pt} 
	\begin{tabularx}{\linewidth}{l l @{\hspace{4mm}} l C C C}
		\toprule
		 & & & \multicolumn{3}{c}{\textbf{Default Setting (mAP)}} \\
      \cline{4-6} \\ [-10pt]
      \# & Keys/Values & Feature head & Full & Rare & N-rare \\
		\midrule
		E1 & Backbone C3 & C4, C5 & 29.59 & 22.41 & 31.73 \\
      E2 & Backbone C4 & C5 & 30.80 & 25.34 & 32.43 \\
      E3 & Backbone C5 & None & \textbf{32.89} & \textbf{27.91} & \textbf{34.38} \\
      F1 & C3 + encoder & None & 30.21 & 25.08 & 31.74 \\
      F2 & C4 + encoder & None & 31.27 & 25.09 & 33.12 \\
      F3 & C5 + encoder & None & 32.69 & 27.38 & 34.27 \\
      G1 & FPN P3 & None & 32.44 & 28.19 & 33.71 \\
      G2 & FPN P4 & None & 32.40 & 28.34 & 33.61 \\
      \midrule
      H1 & Backbone C5 & 1$\times$ Self Attn. & 33.50 & 29.80 & \textbf{34.60}\\
      H2 & Backbone C5 & 2$\times$ Self Attn. & 33.45 & 29.73 & 34.56\\
      H3 & Backbone C5 & 1$\times$ Win. Attn. & 33.54 & 30.17 & 34.55 \\
      H4 & Backbone C5 & 2$\times$ Win. Attn. & \textbf{33.57} & \textbf{30.32} & 34.54 \\
      \bottomrule
	\end{tabularx}
\end{table}

We present the ablation study as a progressive build-up from the baseline model (variant A in \Cref{tab:kv}), which directly feeds the explicit queries into a classifier. All subsequent model variants in this section along with the baseline use a fine-tuned DETR~\cite{detr} with ResNet50 backbone~\cite{resnet}. In \Cref{tab:kv}, we show that introducing cross-attention with encoder features as the keys/values leads to minimal improvement. This indicates that the frozen encoder features may have overfitted to object cues and therefore do not contain orthogonal information beneficial to the understanding of HOIs. The backbone features, on the other hand, contain more general contextual features and result in substantial performance improvement. Nevertheless, there is likely to be a certain degree of overfitting in the backbone, thus warranting investigation into earlier convolutional stages.

\begin{table}[t]\small
	\caption{The mAP ($\times$100) of model variants with different query components and designs on HICO-DET test set. Results are averaged across three runs.}
	\label{tab:query}
	\setlength{\tabcolsep}{4pt} 
	\begin{tabularx}{\linewidth}{@{\extracolsep{\fill}} l l ccccc}
		\toprule
		 & & \multicolumn{2}{c}{\textbf{Modality Fusion}} & \multicolumn{3}{c}{\textbf{Default Setting (mAP)}} \\
      \cline{3-4} \cline{5-7} \\ [-10pt]
      \# & Self-Attn. & Spatial & Content & Full & Rare & N-rare \\
		\midrule
      H3 & Modified & \checkmark & \checkmark & 33.54 & \textbf{30.17} & 34.55 \\
      I1 & Modified & & \checkmark & 33.04 & 28.31 & 34.46 \\
      I2 & None & & \checkmark & 32.60 & 26.79 & 34.34 \\
      I3 & None & \checkmark & & 31.30 & 26.19 & 32.82 \\
      I4 & None & \checkmark & \checkmark & 32.87 & 29.20 & 34.27 \\
      \midrule
      J1 & Vanilla & \checkmark & \checkmark & 33.26 & 29.01 & 34.53 \\
      J2 & Vanilla + pe & \checkmark & \checkmark & \textbf{33.59} & 29.65 & \textbf{34.76} \\
      \bottomrule
	\end{tabularx}
\end{table}

\begin{table}[t]\small
	\caption{The mAP ($\times$100) of model variants with different positional embeddings and number of decoder layers on HICO-DET test set. Results are averaged across three runs.}
	\label{tab:pe}
	\setlength{\tabcolsep}{4pt} 
	\begin{tabularx}{\linewidth}{l l C C C C}
		\toprule
		 & & & \multicolumn{3}{c}{\textbf{Default Setting (mAP)}} \\
      \cline{4-6} \\ [-10pt]
      \# & Positional Embed. & \#Dec. & Full & Rare & N-rare \\
		\midrule
      J2 & None & 1 & 33.59 & 29.65 & 34.76 \\
      K1 & Standard, additive & 1 & 33.43 & \textbf{29.83} & 34.50 \\
      K2 & Standard, concat. & 1 & 33.72 & 29.14 & 35.09 \\
      K3 & Modulated, concat. & 1 & \textbf{33.91} & 29.28 & \textbf{35.29} \\
      \midrule
      L1 & Modulated, concat. & 2 & \textbf{34.18} & \textbf{31.09} & 35.10 \\
      L2 & Modulated, concat. & 3 & 34.03 & 30.18 & \textbf{35.18} \\
      L3 & Modulated, concat. & 4 & 34.05 & 30.44 & 35.12 \\
      \bottomrule
	\end{tabularx}
\end{table}

\begin{table*}[t]\small
   \centering
    \caption{Comparison of detection performance (mAP$\times100$) on the HICO-DET~\cite{hicodet} and V-COCO~\cite{vcoco} test sets. We report results with the common DETR~\cite{detr} detector and ResNet50 backbone, while showing the scalability of our method using the more advanced $\mathcal{H}$-DETR with Swin-L backbone. Best performance in each section is highlighted in bold.}
    \label{tab:results}
    \begin{tabularx}{\linewidth}{@{\extracolsep{\fill}} l l cccccccc}
       \toprule
     & & \multicolumn{6}{c}{\textbf{HICO-DET}} & \multicolumn{2}{c}{\textbf{V-COCO}} \\ [4pt]
     & & \multicolumn{3}{c}{Default Setting} & \multicolumn{3}{c}{Known Objects Setting} & & \\ 
     \cline{3-5}\cline{6-8}\cline{9-10} \\ [-8pt]
       \textbf{Method} & \textbf{Backbone} & Full & Rare & Non-rare & Full & Rare & Non-rare & AP$_{role}^{S1}$ & AP$_{role}^{S2}$ \\
       \midrule
      InteractNet~\cite{interactnet} & ResNet-50-FPN & 9.94 & 7.16 & 10.77 & - & - & - & 40.0 & - \\
      iCAN~\cite{ican} & ResNet-50 & 14.84 & 10.45 & 16.15 & 16.26 & 11.33 & 17.73 & 45.3 & 52.4 \\
      TIN~\cite{tin} & ResNet-50 & 17.03 & 13.42 & 18.11 & 19.17 & 15.51 & 20.26 & 47.8 & 54.2 \\
      VSGNet~\cite{vsgnet} & ResNet-152 & 19.80 & 16.05 & 20.91 & - & - & - & 51.8 & 57.0 \\
      PPDM~\cite{ppdm} & Hourglass-104 & 21.94 & 13.97 & 24.32 & 24.81 & 17.09 & 27.12 & - & - \\
      VCL~\cite{vcl} & ResNet-50 & 23.63 & 17.21 & 25.55 & 25.98 & 19.12 & 28.03 & 48.3 & - \\
      DRG~\cite{drg} & ResNet-50-FPN & 24.53 & 19.47 & 26.04 & 27.98 & 23.11 & 29.43 & 51.0 & - \\
      IDN~\cite{idn} & ResNet-50 & 24.58 & 20.33 & 25.86 & 27.89 & 23.64 & 29.16 & 53.3 & 60.3 \\
      HOTR~\cite{hotr} & ResNet-50 & 25.10 & 17.34 & 27.42 & - & - & - & 55.2 & 64.4 \\
      FCL~\cite{fcl} & ResNet-50 & 25.27 & 20.57 & 26.67 & 27.71 & 22.34 & 28.93 & 52.4 & - \\
      HOI-Trans~\cite{hoitrans} & ResNet-101 & 26.61 & 19.15 & 28.84 & 29.13 & 20.98 & 31.57 & 52.9 & - \\
      AS-Net~\cite{asnet} & ResNet-50 & 28.87 & 24.25 & 30.25 & 31.74 & 27.07 & 33.14 & 53.9 & - \\
      SCG~\cite{scg} & ResNet-50-FPN & 29.26 & 24.61 & 30.65 & 32.87 & 27.89 & 34.35 & 54.2 & 60.9 \\
      QPIC~\cite{qpic} & ResNet-101 & 29.90 & 23.92 & 31.69 & 32.38 & 26.06 & 34.27 & 58.8 & 61.0 \\
      MSTR~\cite{mstr} & ResNet-50 & 31.17 & 25.31 & 32.92 & 34.02 & 28.82 & 35.57 & 62.0 & 65.2 \\
      CDN~\cite{cdn} & ResNet-101 & 32.07 & 27.19 & 33.53 & 34.79 & 29.48 & 36.38 & \textbf{63.9} & 65.9 \\
      UPT~\cite{upt} & ResNet-101-DC5 & 32.62 & 28.62 & 33.81 & 36.08 & 31.41 & 37.47 & 61.3 & \textbf{67.1} \\
      RLIP~\cite{rlip} & ResNet-50 & 32.84 & 26.85 & 34.63 & - & - & - & 61.9 & 64.2 \\
      GEN-VLKT~\cite{gen-vlkt} & ResNet-50 & \textbf{33.75} & \textbf{29.25} & \textbf{35.10} & \textbf{36.78} & \textbf{32.75} & \textbf{37.99} & 62.4 & 64.5 \\
     \midrule
     PViC w/ DETR & ResNet-50 & 34.69 & 32.14 & 35.45 & 38.14 & 35.38 & 38.97 & 62.8 & 67.8 \\
     PViC w/ $\mathcal{H}$-DETR & Swin-L & \textbf{44.32} & \textbf{44.61} & \textbf{44.24} & \textbf{47.81} & \textbf{48.38} & \textbf{47.64} & \textbf{64.1} & \textbf{70.2} \\
       \bottomrule
    \end{tabularx}
\end{table*}

We present the relevant findings in Table~\ref{tab:kv_comp}. For fairness, when using C3 and C4 features (variants E1, E2), we added a feature head equivalent to the missing convolutional stages, and observed that C5 features still yield the best performance. We also explored the composition of the backbone and encoder features illustrated in \Cref{fig:feat_comp}. The results (F variants) show that although the addition of encoder features benefits the lower-level backbone features, it does not introduce orthogonal information to C5 features. In addition, with the G variants, we train a feature pyramid network~\cite{fpn} to propagate the semantics in C5 features to lower levels. The results show that higher-resolution features, albeit helpful for object detection and segmentation~\cite{detr, ta-fcn}, do not benefit the recognition of HOIs. We further investigated adding attention layers to refine the backbone features (H variants) and observed similar performance with self-attention~\cite{xfmer} and window attention~\cite{swint}. Due to the lower complexity, we use window attention in subsequent model variants.
We do not observe significant performance increases with additional attention layers (H2, H4).

Next, we ablate the query components in \Cref{tab:query}. Using variant H3 as the reference, we show the impacts of the object self-attention and modality fusion of object features and the spatial features. Importantly, we show that a vanilla transformer encoder with box positional embeddings achieves comparable performance to the modified encoder in UPT, validating our removal of this custom layer.

Last, we demonstrate the effectiveness of modulated positional embeddings in cross-attention as well as the scaling of decoders in \Cref{tab:pe}. Notably, using the additive positional embeddings (variant K1) does not help the model. This is due to the noise introduced by the two cross-terms in Eq.~\ref{eq:dot_p}. Removing the cross-terms by concatenating the positional embeddings (K2) results in a slight improvement, while the modulated positional embeddings yields additional improvement. Similar to the feature head for keys and values, improvements brought by the decoder saturate after two layers, likely due to the use of frozen features. In summary, we observe a very significant improvement in the rare classes (5 mAP) between our model (variant L1) and the previous state-of-the-art UPT (variant C). This is in line with our understanding that contextual cues introduced with cross-attention greatly benefit the more ambiguous interactions, often the rare ones in the HICO-DET dataset.

\subsection{Comparison with State-of-the-Art Methods}

 We report the performance of our method on HICO-DET~\cite{hicodet} and V-COCO~\cite{vcoco} datasets. For HICO-DET, evaluation is conducted under two different settings. The \textit{default setting} is the primary setting under which different methods are being compared. The criteria for a successful detection extends that of the Pascal VOC challenge~\cite{voc} to bounding box pairs. Specifically, both the human and object boxes need to have an intersection over union (IoU) larger than 0.5 with ground truth for the detected pair to be identified as positive. The \textit{known objects setting} considers the sets of object types of the ground truth pairs in an image to be known, thus automatically removing detections where the object class is outside the set. For V-COCO, there are also two evaluation scenarios, differentiated by the protocol when handling occluded objects. Scenario 1 (S1) requires an empty box prediction for the detection to be considered a match, while scenario 2 (S2) neglects the occluded object and assumes it is always matched.

 \begin{figure*}[t]\captionsetup[subfigure]{font=footnotesize}
   \begin{subfigure}[t]{0.195\linewidth}
     \centering
     \includegraphics[width=\linewidth]{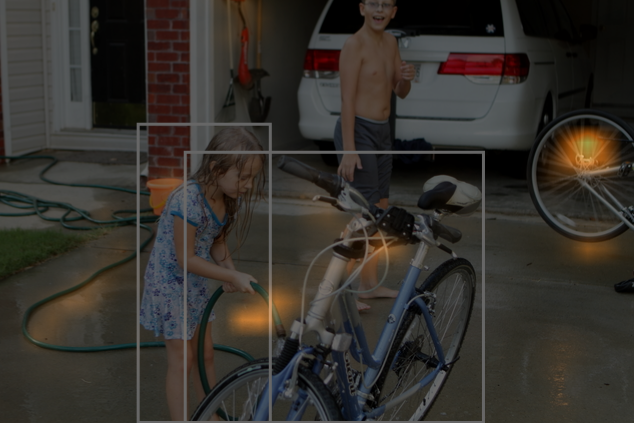}
     \caption{$\sigma(\bk_c\transpose \bq_c)$}
     \label{fig:attn-c}
   \end{subfigure}
   \hfill
   \begin{subfigure}[t]{0.195\linewidth}
     \centering
     \includegraphics[width=\linewidth]{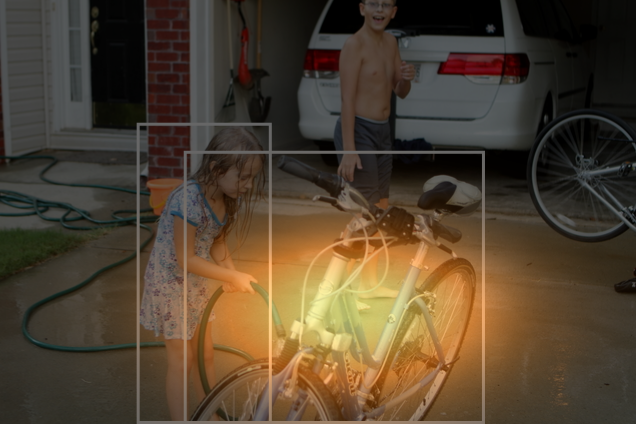}
     \caption{$\sigma(\bk_p\transpose \bq_p)$}
     \label{fig:attn-p}
   \end{subfigure}
   \hfill
   \begin{subfigure}[t]{0.195\linewidth}
     \centering
     \includegraphics[width=\linewidth]{figures/attn_concat.png}
     \caption{$\sigma(\bk_c\transpose \bq_c + \bk_p\transpose \bq_p)$}
     \label{fig:attn-concat}
   \end{subfigure}
   \hfill
   \begin{subfigure}[t]{0.195\linewidth}
     \centering
     \includegraphics[width=\linewidth]{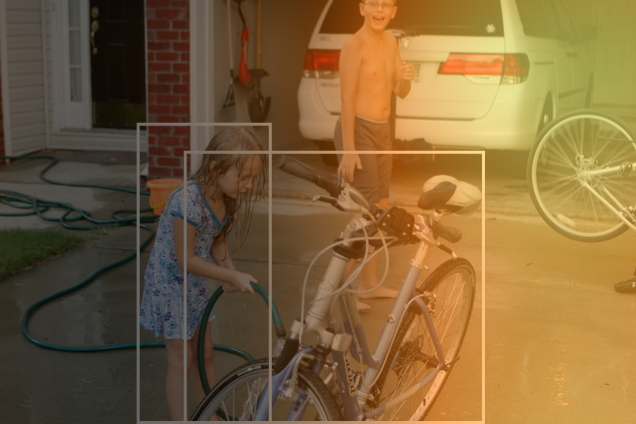}
     \caption{$\sigma(\bk_c\transpose \bq_p)$}
     \label{fig:attn-cp}
   \end{subfigure}
   \hfill
   \begin{subfigure}[t]{0.195\linewidth}
     \centering
     \includegraphics[width=\linewidth]{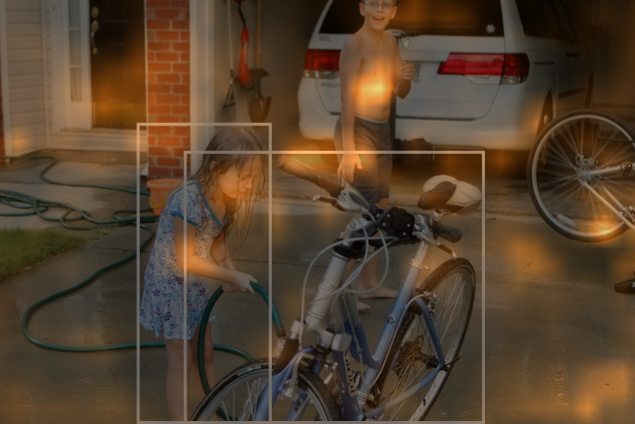}
     \caption{$\sigma(\bk_p\transpose \bq_c)$}
     \label{fig:attn-pc}
   \end{subfigure}
   \vspace{5px}
   \caption{For the image in \Cref{fig:comp_2}, we show visualisations of attention weights computed from the content features~(\subref{fig:attn-c}), positional embeddings~(\subref{fig:attn-p}) and the concatenated formulation~(\subref{fig:attn-concat}). We also show the noisy attention weights computed from the two omitted terms from Eq.~\ref{eq:dot_p} in~(\subref{fig:attn-cp}) and~(\subref{fig:attn-pc}). Here we use $\sigma$ to denote the softmax function and omit the scalar normalisation for brevity of exposition.}
   \label{fig:attn}
\end{figure*}

\begin{figure*}[t]\captionsetup[subfigure]{font=footnotesize}
   \begin{subfigure}[t]{0.195\linewidth}
     \centering
     \includegraphics[width=\linewidth]{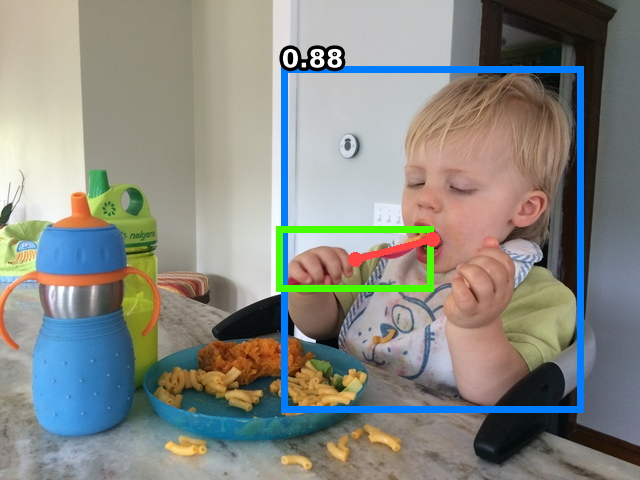}
     \caption{\textit{licking a fork}}
     \label{fig:qual-licking-fork}
   \end{subfigure}
   \hfill
   \begin{subfigure}[t]{0.195\linewidth}
     \centering
     \includegraphics[width=\linewidth]{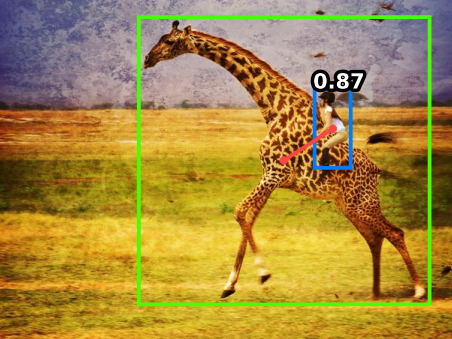}
     \caption{\textit{riding a giraffe}}
     \label{fig:qual-riding-giraffe}
   \end{subfigure}
   \hfill
   \begin{subfigure}[t]{0.195\linewidth}
     \centering
     \includegraphics[width=\linewidth]{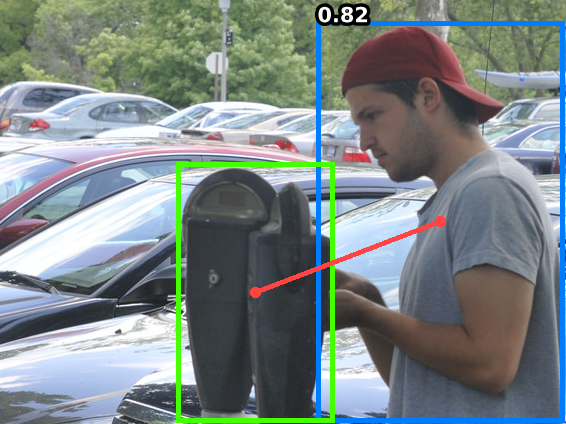}
     \caption{\textit{checking a parking meter}}
     \label{fig:qual-checking-meter}
   \end{subfigure}
   \hfill
   \begin{subfigure}[t]{0.195\linewidth}
      \centering
      \includegraphics[width=\linewidth]{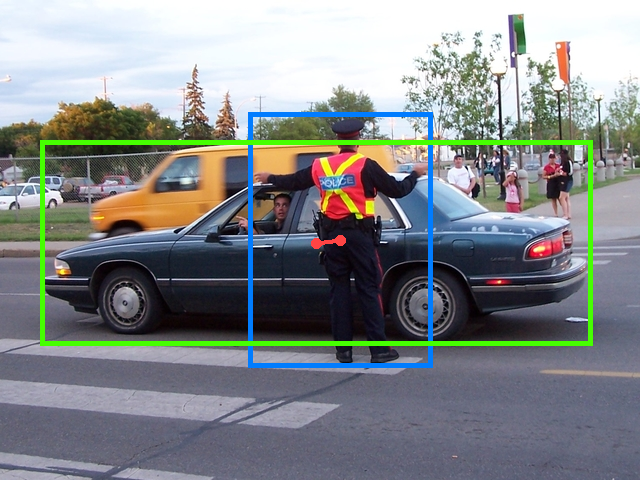}
      \caption{\textit{directing a car}}
      \label{fig:fail-directing-car}
   \end{subfigure}
   \hfill
   \begin{subfigure}[t]{0.195\linewidth}
      \centering
      \includegraphics[width=\linewidth]{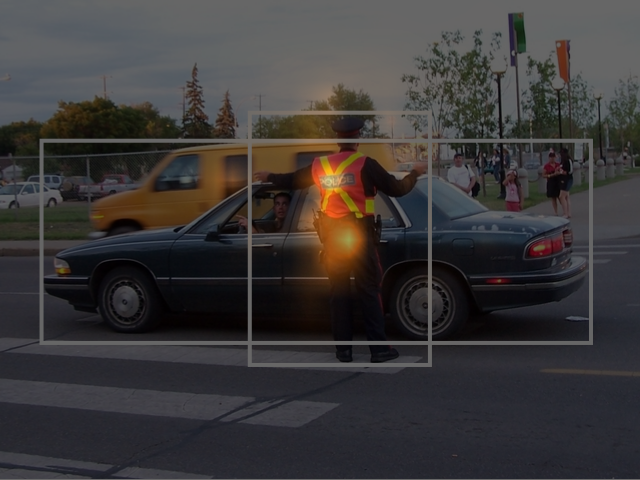}
      \caption{Visualised attention for~(\subref{fig:fail-directing-car})}
      \label{fig:fail-directing-car-attn}
   \end{subfigure}
   \vspace{5px}
   \caption{Qualitative results~(\subref{fig:qual-licking-fork}, \subref{fig:qual-riding-giraffe}, \subref{fig:qual-checking-meter}) and failure case~(\subref{fig:fail-directing-car}) on HICO-DET test set with fine-tuned DETR-R50 as the object detector.}
   \label{fig:qual}
   \vspace*{-3px}
\end{figure*}

We report the performance of our model with two backbones to demonstrate its scalability. For the object detector, we use DETR~\cite{detr} and the most recent $\mathcal{H}$-DETR~\cite{hdetr}, showing the detector-agnostic nature of our approach. As shown in Table~\ref{tab:results}, our method with the ResNet50 already outperforms the previous state-of-the-art two-stage detector UPT by 2.5~mAP, despite it using the heavier ResNet101 and a feature dilation. Furthermore, compared against GEN-VLKT, which distils features from a vision and language model trained on millions of images (CLIP~\cite{clip}), our method achieves higher performance with the same ResNet50 backbone. With a stronger detector and backbone, \ie, $\mathcal{H}$-DETR, our method receives significant performance boost. This highlights one of the great advantages of two-stage detectors, that they can directly benefit from independent advances in object detection.

\subsection{Positional Embeddings in Cross-Attention}

To elucidate the mechanism of positional embeddings, we separate the attention weights for each term in Eq.~\ref{eq:dot_p} and visualise them in \Cref{fig:attn}. Starting with the content features of the keys and queries ($\bk_c\transpose \bq_c$), we show in \Cref{fig:attn-c} that this term only accounts for the visual similarity. Consequently, a distant object of a relevant class, i.e. the bike in the background, receives a substantial amount of attention. This can be corrected by the similarity term between positional embeddings, which places a spatial bias on locations of the human object pair as depicted in \Cref{fig:attn-p}. When the positional embeddings are concatenated to the content features, the resultant attention weights become the sum of the two terms. As shown in \Cref{fig:attn-concat}, combination of these two terms results in high attention weights on the water hose, which is the key to recognising the interaction \textit{washing a bike}. In addition, we show in Figures~\ref{fig:attn-cp} and~\ref{fig:attn-pc} that the two omitted cross-terms from Eq.~\ref{eq:dot_p} mostly generate noise, as the content features and positional embeddings are from very different feature spaces, justifying their removal.

\subsection{Qualitative Results and Limitations}

We show additional qualitative results in this section. In particular, our model performs well on several interactions with little training data, such as \textit{licking a fork} (six training examples,~\ref{fig:qual-licking-fork}), \textit{riding a giraffe} (two training examples,~\ref{fig:qual-riding-giraffe}) and \textit{checking a parking meter} (36 training examples,~\ref{fig:qual-checking-meter}). We also show an example of missed detections in \Cref{fig:fail-directing-car}, due to a severe lack of training examples, one in this case. In addition, as there are other interaction classes involving the same predicate \textit{directing} but with different objects and backgrounds, the model tends to fit towards other classes. Consequently, the model cannot locate the relevant visual context~(\Cref{fig:fail-directing-car-attn}), hand gesture in this case.

\section{Conclusion}

In this paper we analysed the visual features used in existing two-stage HOI detectors and concluded that their major weakness was a lack of relevant contextual information, since they were specialised to the localisation task. As such, we proposed an improved design by re-introducing image features into the human--object pair representation via cross-attention. To this end, we performed extensive experiments on the choices of keys/values and introduced box pair positional embeddings as spatial guidance, and visualised the impacts of the attention mechanism. Compared to previous two-stage approaches, we streamlined and simplified the architecture, reducing the need for custom components. Our method achieves state-of-the-art performance on the relevant benchmarks, with particular improvements where fine-grained visual features, like human pose, and additional context, like another object involved in the interaction, are relevant to the classification.

\newpage

{\small
\bibliographystyle{ieee_fullname}
\bibliography{egbib}
}

\newpage
\appendix

\section{Bounding Box Pair Positional Embeddings}

We provide more mathematical details on the positional embeddings for bounding box pairs used in cross-attention. Let us first revisit the notations from the main paper. We define sinusoidal embedding of a scalar as $\boldsymbol{\phi}: \mathbb{R} \rightarrow \mathbb{R}^d$,
\begin{equation}
   \boldsymbol{\phi}(x)_{2i} = \sin\left(\frac{x}{\tau^{2i/d}}\right),\; \boldsymbol{\phi}(x)_{2i - 1} = \cos\left(\frac{x}{\tau^{2i/d}}\right),
\end{equation}
where $i=1,\ldots,d/2$ and $\tau$ is a temperature parameter we set as $20$ following previous practice~\cite{dab-detr}. With box width and height as modulation, the positional embeddings for one bounding box are as follows
\begin{equation}
   \text{PE}(x, y, w, h) = \left[ \boldsymbol{\phi}(y) \frac{h_{ref}}{h}, \boldsymbol{\phi}(x) \frac{w_{ref}}{w} \right] \in \mathbb{R}^{2d},
\end{equation}
where $w_{ref}$, $h_{ref}$ are reference values learned from box appearance features $\boldf$ as follows
\begin{equation}
   w_\text{ref}, h_\text{ref} = \sigma(\text{MLP}(\boldf)),
\end{equation}
where $\sigma$ is the sigmoid function. As such, the positional embeddings for a human--object pair ($\bb_h, \bb_o \in \mathbb{R}^4$) are defined by concatenating the positional embeddings of the two boxes,
\begin{equation}
   \bq_p = \left[ \bq_p^h, \bq_p^o \right] = \left[ \text{PE}(\bb_h), \text{PE}(\bb_o) \right] \in \mathbb{R}^{4d}.
\end{equation}
Denote the positional embeddings of an image patch with normalised spatial indices ($i, j$) by
\begin{equation}
   \bk_p = [\boldsymbol{\phi}(j), \boldsymbol{\phi}(i)] \in \mathbb{R}^{2d}.
\end{equation}
Assuming the number of heads is one, the dot-product attention weights between positional embeddings are computed as
\begin{equation}
   (W_k \bk_p)^T (W_p \bq_p ) = \bk_p^T W_k^T W_p \bq_p,
   \label{eq:dot-prod}
\end{equation}
where $W_k \in \mathbb{R}^{2d\times 2d}, W_p \in \mathbb{R}^{2d \times 4d}$ are weight matrices associated with the linear transformations applied to the positional embeddings. In particular, matrix $W_p$ can be partitioned into $\left[ W_p^h, W_p^o \right]$, therefore decomposing the linear transformation on query (human--object pair) positional embeddings as follows
\begin{equation}
   W_p \bq_p = W_p^h \bq_p^h + W_p^o \bq_p^o.
\end{equation}
For brevity of exposition, let us now assume that weight matrices $W_k, W_p^h, W_p^o \in \mathbb{R}^{2d\times 2d}$ are identity matrices. This simplifies the dot-product attention weights between positional embeddings in Eq.~\ref{eq:dot-prod} as follows
\begin{equation}
   \bk_p^T \bq_p^h + \bk_p^T \bq_p^o,
\end{equation}
demonstrating that the concatenation of two modulated box positional embeddings results in a weighted sum of the pre-normalised attention weights.

\section{Multiscale Features as Keys/Values}

Deformable DETR~\cite{deform-detr} introduced a simple way to exploit the multiscale structure of convolutional features via deformable attention. Specifically, for each query, a set of offsets with respect to a reference point are predicted to obtain a small number of keys and values. The attention operation is thus reduced to a sparse variant between a query and its corresponding subset of keys and values. As a result of the sparsity, it becomes affordable to extend the cross-attention source from a single feature level to a feature pyramid, where keys/values across different levels are concatenated.

\begin{figure}
   \begin{subfigure}[t]{0.326\linewidth}
      \centering
      \includegraphics[width=\linewidth]{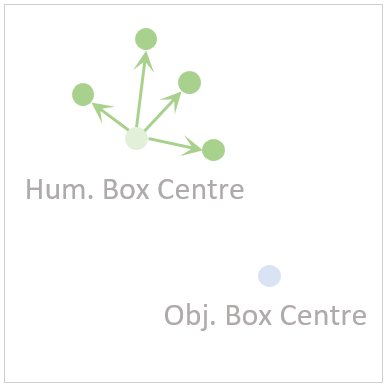}
      \caption{Single ref.}
      \label{fig:defm_ref_1}
   \end{subfigure}
   \hfill
   \begin{subfigure}[t]{0.326\linewidth}
      \centering
      \includegraphics[width=\linewidth]{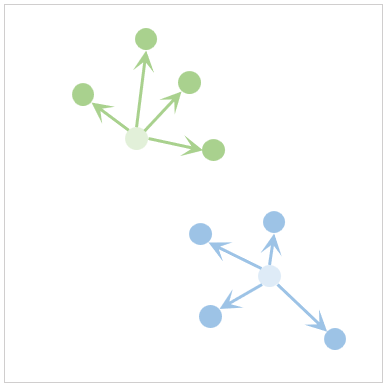}
      \caption{Dual ref.}
      \label{fig:defm_ref_2}
   \end{subfigure}
   \hfill
   \begin{subfigure}[t]{0.326\linewidth}
      \centering
      \includegraphics[width=\linewidth]{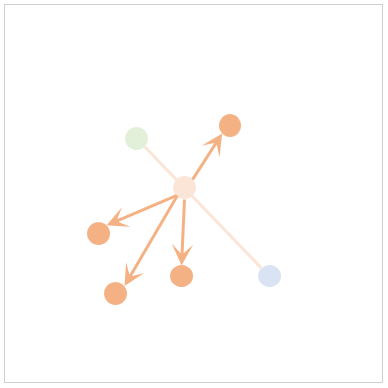}
      \caption{Single ref.}
      \label{fig:defm_ref_3}
   \end{subfigure}
   \vspace{1px}
   \caption{Reference point designs for human--object pairs. (\subref{fig:defm_ref_1}) Human box centre as the single reference point. (\subref{fig:defm_ref_2}) Human and object box centres as dual reference points. (\subref{fig:defm_ref_3}) Interpolated point as the single reference point.}
   \label{fig:defm_ref}
\end{figure}

In the case of two-stage HOI detection, we follow the practice in Deformable DETR to construct multiscale features denoted by $\{C_3, C_4, C_5, C_6\}$, where $C_3, C_4, C_5$ are extracted directly from the backbone and $C_6$ is obtained from $C_5$ by applying a $3\times 3$ convolution with stride $2$. Four sets of keys/values are sampled for each feature level per query. For the reference points, Deformable DETR uses bounding box centres predicted from the query representations. In our two-stage method, due to the availability of bounding boxes, there is no need to predict the reference points. As such, we focus on designing reference points for human--object pairs and explore three variants depicted in~\Cref{fig:defm_ref}.

As humans play a centric role in HOIs, we start with a simple variant with human box centres as the reference point for each query~(\Cref{fig:defm_ref_1}). Naturally, this can be extended to dual reference points to also include the object box centre~(\Cref{fig:defm_ref_2}). Last, we experiment with a variant where the reference point is a convex combination (linear interpolation) of the human and object box centres~(\Cref{fig:defm_ref_3}). Formally, denote the box centres by $x, y \in \mathbb{R}^2$. The reference point is computed as $\beta x + (1 - \beta) y$, where the scaling factor $\beta \in [0, 1]$ is predicted from the query representation and normalised with sigmoid.

\begin{table}[t]\small
	\caption{The mAP ($\times$100) of model variants with different reference point designs for deformable attention. Results are averaged across three runs.}
	\label{tab:ms}
	\setlength{\tabcolsep}{4pt} 
	\begin{tabularx}{\linewidth}{l l C C C C}
		\toprule
		 & & & \multicolumn{3}{c}{\textbf{Default Setting (mAP)}} \\
      \cline{4-6} \\ [-10pt]
      \# & Ref. points & \#Keys & Full & Rare & N-rare \\
		\midrule
      L1 & N/A & N/A & \textbf{34.18} & \textbf{31.09} & \textbf{35.10} \\
      \midrule
      M1 & single (hum.) & 4 & 33.55 & 29.70 & 34.70 \\
      M2 & dual & 8 & 33.59 & 30.31 & 34.57 \\
      M3 & dual & 4 & 33.55 & 29.88 & 34.65 \\
      M4 & single (interp.) & 4 & 33.40 & 29.58 & 34.54 \\
      \bottomrule
	\end{tabularx}
\end{table}

\begin{figure}[t]\captionsetup[subfigure]{font=footnotesize}
   \begin{subfigure}[t]{0.495\linewidth}
     \centering
     \includegraphics[width=\linewidth]{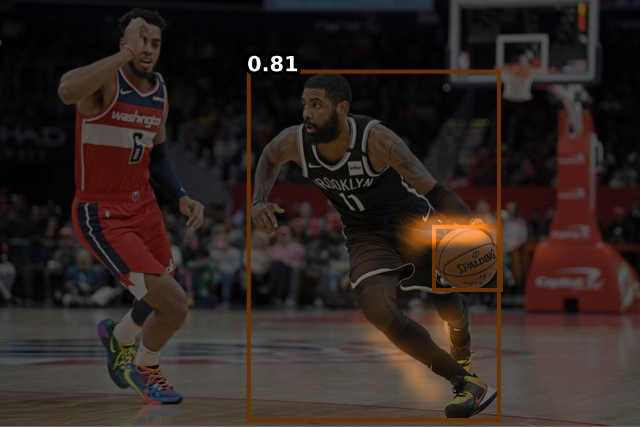}
     \caption{\textit{dribbling a sports ball}}
     \label{fig:qual-dribbling-sportsball}
   \end{subfigure}
   \hfill%
   \begin{subfigure}[t]{0.495\linewidth}
     \centering
     \includegraphics[width=\linewidth]{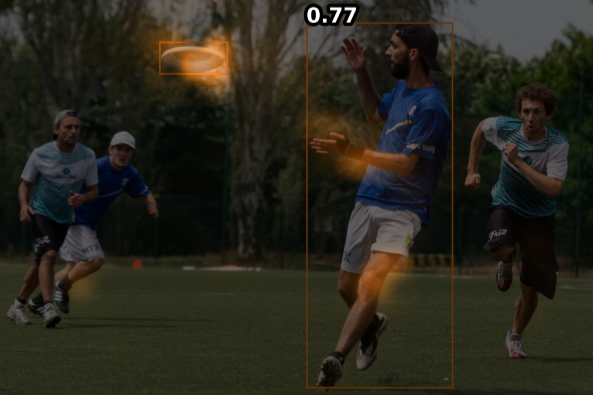}
     \caption{\textit{catching a frisbee}}
     \label{fig:qual-catching-frisbee}
   \end{subfigure}

   \begin{subfigure}[t]{\linewidth}
      \centering
      \includegraphics[width=.495\linewidth]{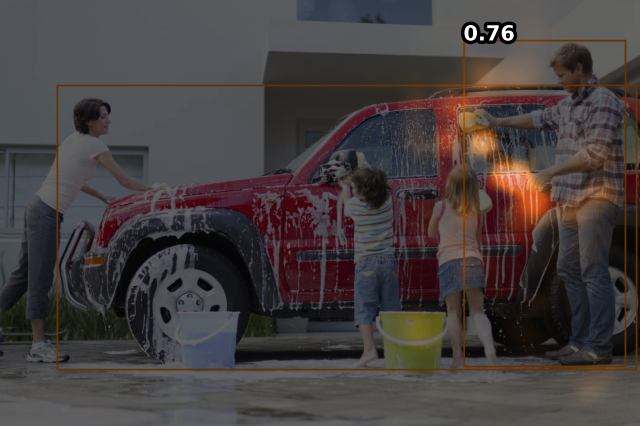}
      \hfill%
      \includegraphics[width=.495\linewidth]{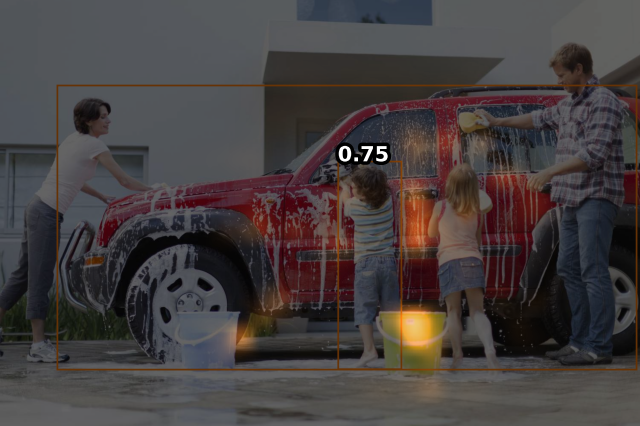}
      \vfill%
      \includegraphics[width=.495\linewidth]{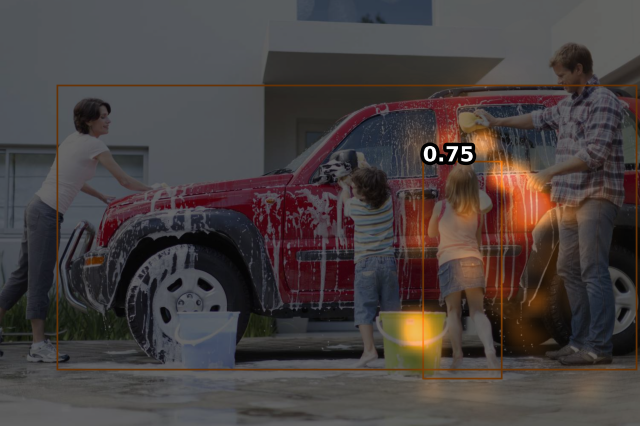}
      \hfill%
      \includegraphics[width=.495\linewidth]{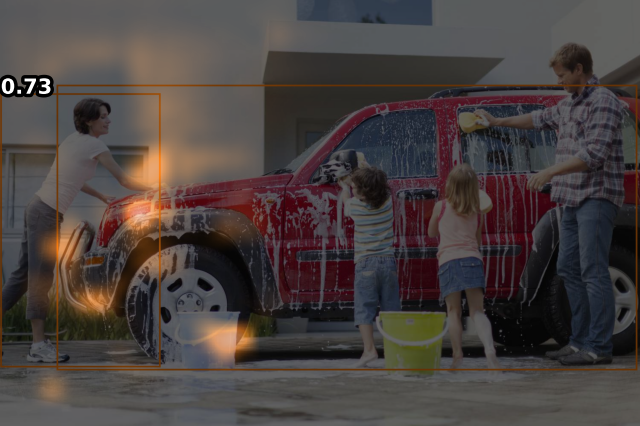}
      \caption{\textit{washing a car}}
      \label{fig:qual-washing-car}
    \end{subfigure}
   
   \vspace{5px}
   \caption{Predicted human--object interactions and visualised attention weights on data in the wild.}
   \label{fig:add-qual}
\end{figure}

We compare the performance against the L1 variant, which uses the C5 features as keys/values and employs a vanilla decoder with box pair positional embeddings introduced in the main paper. For fair comparison, we use one deformable encoder layer to refine the multiscale features and two deformable decoder layers, a similar setup to the L1 variant. As shown in~\Cref{tab:ms}, we observed insignificant performance differences amongst the M variants with multiscale features, while their performance in general lacks behind the single-scale variant L1.

We believe the inferior performance of multiscale deformable attention is likely due to the visual complexity of human--object interactions. In particular, the role of reference points is somewhat similar to the box pair positional embeddings. Although the offsets with respect to the reference points are dynamically predicted from the query representation, they tend not to have large values. Therefore, they act as a form of inductive bias to encourage high attention weights on keys/values (image patches) closer to the reference point, analogous to the $\bk_p^T \bq_p$ term in the vanilla decoder. On the other hand, a weakness of deformable attention is that, it does not have a mechanism to encourage attention based on visual similarity,~\ie the $\bk_c^T \bq_c$ term in the vanilla decoder. As we have demonstrated in the main paper, the recognition of HOIs requires more complex visual context, compared to object detection. Therefore, the offsets predicted from the query features are not sufficient to locate the relevant context.

\section{Demonstration on Data in The Wild}

For more evidence on the two types of visual context exploited by our model, we provide additional qualitative results on data in the wild, with cross-attention weights overlaid. As shown in Figures~\ref{fig:qual-dribbling-sportsball} and~\ref{fig:qual-catching-frisbee}, the model extracts contextual features from image regions containing relevant human body parts, and successfully predicts the correct interactions with high scores. In~\Cref{fig:qual-washing-car}, we highlight the other type of context, \ie another involved object. Notably, three out of the four human--object pairs place high attention weights on the water buckets, which are indicators of the corresponding interaction.

\section{Pipeline}

\begin{figure}
    \centering
    \includegraphics[width=.9\linewidth]{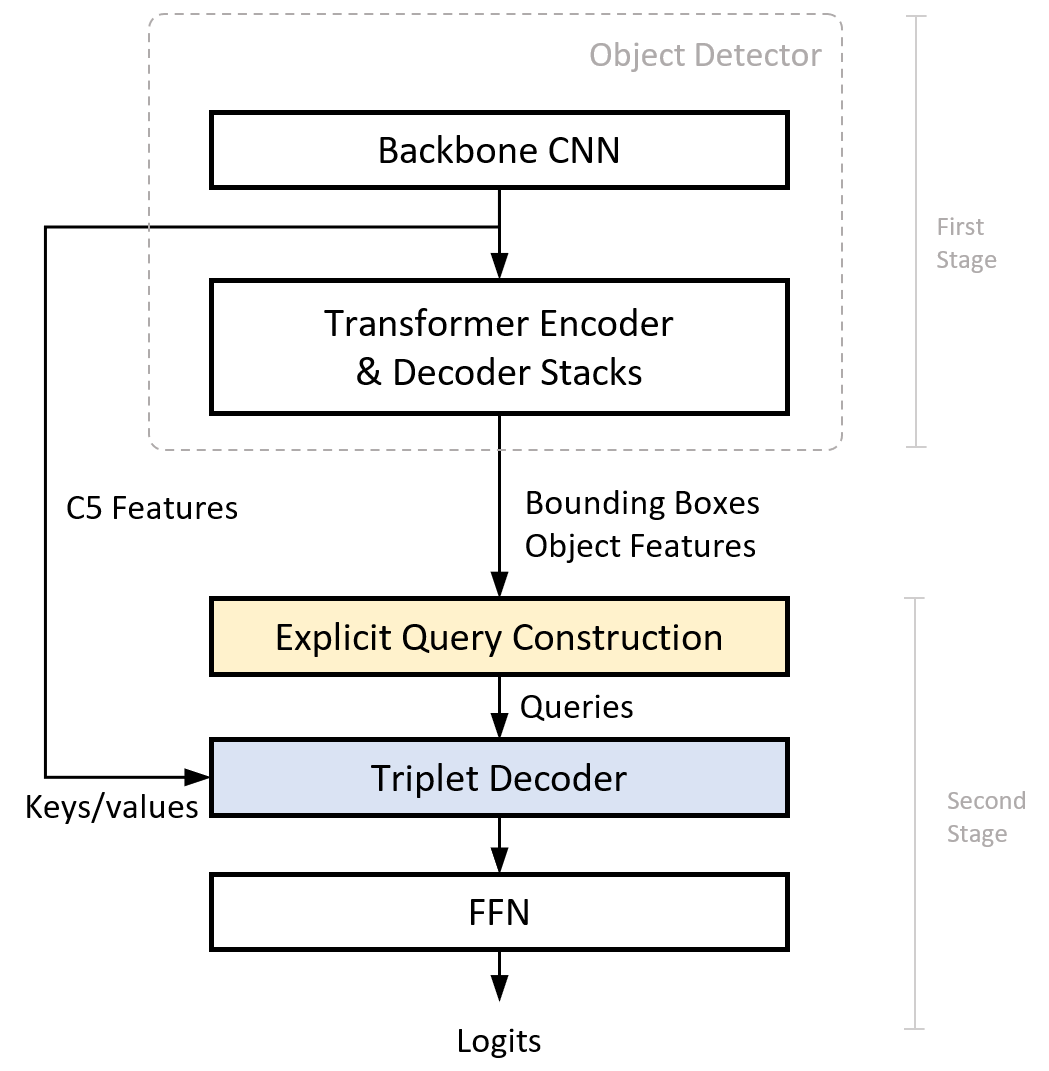}
    \caption{Illustration of the overall pipeline.}
    \label{fig:pipeline}
\end{figure}

For better clarity, we attach an illustration of the entire pipeline, as shown in Figure~\ref{fig:pipeline}. Due to the popularity of the DETR framework, the first stage is depicted as a transformer-based object detector. But the method itself is detector-agnostic.

\section{Advanced Variants of DETR}

To demonstrate the detector-agnostic nature of our method,
we fine-tuned the recent state-of-the-art object detector $\mathcal{H}$-DETR~\cite{hdetr}
and reported the performance of our method with it, as shown in Table~\ref{tab:results}.
However, we would like to point out that the performance of our method with $\mathcal{H}$-DETR-R50
was surprisingly lower than that with DETR-R50,
although $\mathcal{H}$-DETR-R50 outperforms DETR-R50 significantly in terms of object detection mAP on HICO-DET~\cite{hicodet}.
To investigate this issue,
we first made the observation that $\mathcal{H}$-DETR was trained using a multi-label classification objective,
that is, the scores are individual normalised using the Sigmoid function as opposed to Softmax.
As a result, the predicted object detection scores tend to be lower, thus less over-confident.
To this end, we increased the value of hyper-parameter $\lambda$ from Eq.~\ref{eq:score} to $0.37$.
Nevertheless, this only resulted in marginal improvement.

In addition, $\mathcal{H}$-DETR employs deformable attention~\cite{deform-detr} and utilises the multi-scale feature.
Our method, on the other hand, uses a single-scale feature in the spatially-guided cross-attention.
Therefore, it is likely that the different levels of the feature maps learned to extract different information,
Unfortunately, our attempts in exploiting such multi-scale features did not lead to concrete improvements.
As such, we leave this problem to potential future work.

\end{document}